\documentclass[10pt,twocolumn,letterpaper]{article}

\usepackage[pagenumbers]{cvpr}      %
\usepackage[dvipsnames]{xcolor}
\usepackage[linesnumbered,ruled,lined]{algorithm2e}

\usepackage{graphicx}
\usepackage{url}
\usepackage{animate}
\usepackage{multirow}
\newcommand\blfootnote[1]{\begingroup\renewcommand\thefootnote{}\footnote{#1}\addtocounter{footnote}{-1}\endgroup}

\newcommand{\myPara}[1]{\noindent\textbf{#1}}
\usepackage{amsmath}

\definecolor{cvprblue}{rgb}{0.21,0.49,0.74}
\usepackage[pagebackref,breaklinks,colorlinks,citecolor=cvprblue]{hyperref}

\def\paperID{6182} %
\def\confName{CVPR}
\def\confYear{2024}

\title{VideoSwap: Customized Video Subject Swapping with \\ Interactive Semantic Point Correspondence}

\author{Yuchao Gu$^{1,2}$,\; Yipin Zhou$^{2}$,\; Bichen Wu$^{2}$,\; Licheng Yu$^{2}$,\; Jia-Wei Liu$^{1}$,\; Rui Zhao$^{1}$,\; Jay Zhangjie Wu$^{1}$,\; \\ David Junhao Zhang$^{1}$,\; Mike Zheng Shou$^{1}$\thanks{Corresponding Author.},\; Kevin Tang$^{2}$\; \\ \\
$^1$Show Lab, National University of Singapore\quad $^2$GenAI, Meta \\
\url{https://videoswap.github.io/}
}

\newcommand{\secref}[1]{Sec.~\ref{#1}}
\newcommand{\tabref}[1]{Table.~\ref{#1}}
\newcommand{\figref}[1]{Fig.~\ref{#1}}
\newcommand{\algref}[1]{Algorithm.~\ref{#1}}

\newcommand{\teaser}{
  \vspace{-3em}
  \begin{center}
\animategraphics[width=\linewidth,loop]{8}{imgs/teaser1/}{00001}{00016}\end{center}
  \vspace{-.26in}
  \captionof{figure}{Customized video subject swapping results with VideoSwap. VideoSwap supports shape change in the swapped results while aligning with the source motion trajectory. The swapped target can be either a predefined concept from a pretrained model (\eg, {\color{ForestGreen}\textit{helicopter}}) or a customized concept (denoted by {\color{Fuchsia}$V^*$}). Previous methods based on implicit motion encoding and dense correspondence do not perform well in subject swapping with shape changes. We encourage readers to \textcolor{magenta}{click and play} the video clips in this figure using Adobe Acrobat.}
  \vspace{1.5em}
  \label{fig:teaser1}
}

\begin{document}

\twocolumn[{%
\vspace{-2em}
\maketitle%
\teaser%
}]\blfootnote{$^*$Corresponding Author}

\begin{abstract}
Current diffusion-based video editing primarily focuses on structure-preserved editing by utilizing various dense correspondences to ensure temporal consistency and motion alignment. 
However, these approaches are often ineffective when the target edit involves a shape change.
To embark on video editing with shape change, we explore customized video subject swapping in this work, where we aim to replace the main subject in a source video with a target subject having a distinct identity and potentially different shape.
In contrast to previous methods that rely on dense correspondences, we introduce the VideoSwap framework that exploits semantic point correspondences, inspired by our observation that only a small number of semantic points are necessary to align the subject's motion trajectory and modify its shape. We also introduce various user-point interactions (\eg, removing points and dragging points) to address various semantic point correspondence.
Extensive experiments demonstrate state-of-the-art video subject swapping results across a variety
of real-world videos.
\end{abstract}

\begin{figure*}[!tb]
    \centering
    \includegraphics[width=0.93\linewidth]{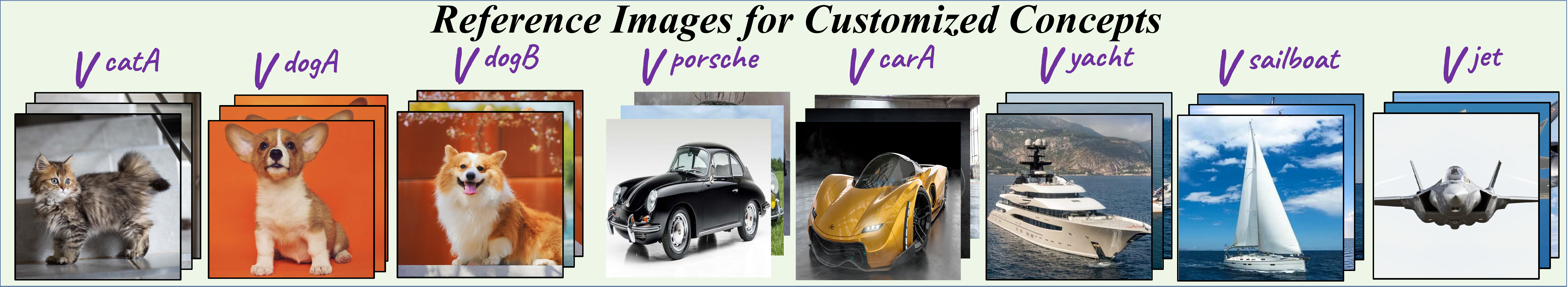}
    \animategraphics[width=0.93\linewidth,loop]{8}{imgs/teaser2_all/animal/}{00001}{00016}
    \animategraphics[width=0.93\linewidth,loop]{8}{imgs/teaser2_all/object/}{00001}{00016}
    \vspace{-.13in}
    \caption{Customized video subject swapping results of VideoSwap on various concepts. The swapping target can either be a predefined concept in the pretrained model (\eg, {\color{ForestGreen}\textit{helicopter}}) or a customized concept created by ED-LoRA~\cite{gu2023mix} (denoted as {\color{Fuchsia}$V^*$}).
    We encourage readers to \textcolor{magenta}{click and play} the video clips in this figure using Adobe Acrobat. \textbf{For legal issues, we cannot display the human swap results.}}
    \label{fig:teaser2}
    \vspace{-.13in}
\end{figure*}

\section{Introduction}

Diffusion-based video editing~\cite{chai2023stablevideo,zhao2023controlvideo, wu2023tune, liew2023magicedit, qi2023fatezero, molad2023dreamix, esser2023structure,wu2023cvpr} is an emerging field that harnesses the capabilities of pretrained text-to-image/video diffusion models~\cite{rombach2022high, guo2023animatediff, ho2022imagen, singer2022make} to facilitate a range of video editing tasks, including style change and subject/background swapping.
The main challenge in video editing lies in how to extract motion from the source video and transfer it to the edited video while ensuring temporal consistency. 
Pioneer Tune-A-Video~\cite{wu2023tune} implicitly encodes source motion in the diffusion model weights by tuning from the source video. While it demonstrates versatile applications for video editing, the temporal consistency is far from satisfactory. Subsequent works make use of various dense correspondences extracted from the source video, including attention maps~\cite{qi2023fatezero,liu2023video}, edge/depth maps~\cite{yang2023rerender,zhao2023controlvideo, liew2023magicedit, esser2023structure}, optical flows~\cite{yang2023rerender}, and deformation fields~\cite{chai2023stablevideo,couairon2023videdit} for video editing. While achieving better temporal consistency, dense correspondence imposes strict shape constraints on the target edit, which makes it ineffective for video editing with shape changes.

To embark on video editing with shape change, we delve into a challenging task: customized video subject swapping. Unlike the conventional video subject swapping addressed in previous works~\cite{yang2023rerender, chai2023stablevideo, zhao2023controlvideo}, where the swapped subject conforms to the shape of source subject, customized subjects have a clearly defined identity in terms of both appearance and shape. These distinctive characteristics should be preserved in the target edit.
Therefore, previous structure-preserved video editing methods  are often ineffective for this problem, as shown in \figref{fig:teaser1}(b).

To address customized video subject swapping, our primary insight is that \textbf{\textit{the subject's motion trajectory can be effectively described using a small number of semantic points}}. As shown in \figref{fig:teaser1}(a), the motion trajectory of an airplane can be depicted by semantic points located at its wings, nose, and tail.
This insight naturally leads us to the following question: Can we utilize these semantic points as correspondences to align the motion trajectory while relaxing the shape constraints in video editing?
To answer this question, we conduct a toy experiments and observe that it is possible to learn semantic point correspondence for a specific video subject using just a small number of source video frames. Users can interact with learned semantic points to generate unseen poses or modify the shape of the video subject.
These observations suggest the potential for integrating semantic point correspondence into video editing, provided that we can obtain an accurate semantic point sequence for the target edit.

To unleash the potential of semantic point correspondence, we introduce the VideoSwap framework for customized video subject swapping, which comprises the following primary designs:
1) Integrating the motion layer into the image diffusion model to ensure essential temporal consistency.
2) Registering semantic points on the source video and utilizing them to transfer the motion trajectory of source subject to the target edit.
3) Introducing user-point interactions (\eg, removing or dragging points) for various semantic point correspondence.

Our contributions are summarized as follows:
\begin{itemize}
\item Empirical observations that reveal the potential of semantic point correspondence for aligning motion trajectories and changing shapes in video editing.
\item The VideoSwap framework, which minimizes user intervention while unleashing the potential of semantic point correspondence in customized video subject swapping.
\item State-of-the-art results in customized video subject swapping, as demonstrated in \figref{fig:teaser2}.
\end{itemize}
\section{Related Work}

\subsection{Diffusion-Based Video Editing}

\myPara{Structure-Preserved Video Editing.}
FateZero~\cite{qi2023fatezero} and Video-P2P~\cite{liu2023video} extract cross- and self-attention from the source video to control spatial layout.
To achieve stricter alignment of temporal consistency with the source video, Rerender-A-Video~\cite{yang2023rerender}, Gen-1~\cite{esser2023structure}, ControlVideo~\cite{zhao2023controlvideo}, and TokenFlow~\cite{geyer2023tokenflow} extract and align optical flow, depth/edge maps, and nearest-neighbour field from the source video respectively, resulting in improved temporal consistency. StableVideo~\cite{chai2023stablevideo}, VidEdit~\cite{couairon2023videdit}, and CoDEF~\cite{ouyang2023codef} learn the canonical space for editing following the Layered Neural Atlas~\cite{kasten2021layered} or the deformation field in Dynamic Nerf~\cite{mildenhall2021nerf,pumarola2021d}.
While achieving promising results in structure-preserved video editing, the above methods based on various dense correspondence are not suitable for handling subject swapping involving shape changes.

\myPara{Video Editing with Shape Change.}
Tune-A-Video (TAV)~\cite{wu2023tune} and FateZero~\cite{qi2023fatezero} with TAV checkpoint can be utilized for video editing with shape change, as they implicitly encode the motion in model weights through tuning on the source video. However, they suffer from structure and appearance leakage due to model tuning.
Shape-aware editing~\cite{lee2023shape} is built on the Layered Neural Atlas~\cite{kasten2021layered}, employing semantic correspondences to estimate shape deformation and warp the atlas. Nevertheless, texture warping will lead to unrealistic textures.

\myPara{Video Diffusion Models.} Previous video editing primarily relies on the text-to-image diffusion model~\cite{rombach2022high}. However, recent advancements are occurring in video foundation models~\cite{guo2023animatediff, blattmann2023align, wang2023modelscope, zhang2023show}. In this work, we add a motion layer~\cite{guo2023animatediff} to the image diffusion model to provide essential temporal consistency for video editing, and we focus on exploiting semantic point correspondence to align the motion trajectory of the video subject.

\subsection{Point Correspondence}

\myPara{Point Correspondence in Diffusion Models.}
DIFT~\cite{tang2023emergent} initially uncovers robust semantic point correspondences in diffusion models. Building upon the observation in DIFT, subsequent works, DragDiffusion~\cite{shi2023dragdiffusion} and DragonDiffusion~\cite{mou2023dragondiffusion} extend DragGAN~\cite{pan2023drag} to support interactive point-based image editing in diffusion models. However, point correspondences and interactive drag-based editing are seldom investigated for video editing.

\myPara{Tracking Any Point in Video.}
TAP-Vid~\cite{doersch2022tap} first introduces the problem of tracking any point (TAP). Unlike optical flow estimation, TAP requires the establishment of long-range correspondence for all points in a video. In our work, we use TAP to reduce human intervention in annotating subject keypoints and acquire long-range motion estimation.
Although several works~\cite{doersch2022tap,wang2023tracking, karaev2023cotracker,doersch2023tapir} address the TAP problem, we choose to employ Co-Tracker~\cite{karaev2023cotracker}, which is the most efficient solution available.

\subsection{Concept Customization}
Concept customization is mainly categorized into tuning-based approaches~\cite{gal2022image, voynov2023p+, ruiz2022dreambooth, kumari2022multi, gu2023mix} and tuning-free solutions~\cite{shi2023instantbooth, ye2023ip, jia2023taming, ruiz2023hyperdreambooth, xiao2023fastcomposer}. Tuning-free solutions are fast but typically adhere closely to the provided reference image and lack variation. On the other hand, tuning-based solutions can leverage multi-view images to ensure variation in the given concepts and maintain the same inference behavior to pretrained diffusion models.
In this paper, we employ the tuning-based ED-LoRA~\cite{gu2023mix} for encoding subject identity.

In addition to image customization for generation (\ie, noise to image), several works employ customization techniques for subject-driven image editing (\ie, image to image). CustomEdit~\cite{choi2023custom} and Photoswap~\cite{gu2023photoswap} propose the invert concept identity to text token and utilize attention swapping to preserve the layout and pose of the source image. While these methods achieve promising swapping results, the injection of attention maps tends to constrain the shape and leak color information to the target swapped result, as observed in DreamEdit~\cite{li2023dreamedit}. In contrast to them, we only align the semantic points' correspondence with the source subject and thus relax the shape constraints, better revealing the concept identity.
\begin{figure}[!tb]
    \centering
\animategraphics[width=0.95\linewidth,loop]{8}{imgs/observation/}{00001}{00008}
    \vspace{-.15in}
    \caption{Toy experiment exploring semantic point correspondence.
    We encourage readers to \textcolor{magenta}{click and play} the video clips in this figure using Adobe Acrobat.}
    \vspace{-.1in}\label{fig:observation}
\end{figure}

\section{VideoSwap}
In this section, we start by presenting a toy experiment to illustrate our motivation to explore semantic point correspondence in \secref{sec:motivation}. Subsequently, we offer an overview of the VideoSwap pipeline for customized video subject swapping in \secref{sec:overview}. Following that, we explain the process of injecting semantic point correspondence in \secref{sec:source_point_extract} to \secref{sec:point_propagation}.

\subsection{Motivation}
\label{sec:motivation}
Dense correspondences explored in previous video editing methods~\cite{chai2023stablevideo, yang2023rerender, geyer2023tokenflow,esser2023structure,ouyang2023codef} restrict the subject's shape change in the edited video. Therefore, our goal is to find a more flexible correspondence that can transfer the source subject's motion trajectory without imposing strict shape constraints.
Motivated by this, we investigate sparse semantic points as correspondences. Unlike dense correspondences such as depth, edge, and optical flow, which are low-level cues shared across all video subjects, semantic points vary with different open-world concepts. Therefore, it is not feasible to train a general condition model for injecting semantic point correspondence. Instead, the question we aim to address in this section is, \textit{\textbf{is it possible to learn semantic point control for a specific source video subject using only a small number of source video frames}}?

\myPara{Toy Experiment Setting.}
To address the above question, we perform a toy experiment. Firstly, for a given video, we manually define a set of semantic points. Next, we annotate the same set of points on eight frames of this video. Finally, we train a T2I-Adapter~\cite{mou2023t2i} on these data pairs, as illustrated in \figref{fig:observation}(a, b), to determine whether these semantic points can be used to control the source video subject.

\begin{figure}[!tb]
    \centering
\includegraphics[width=0.99\linewidth]{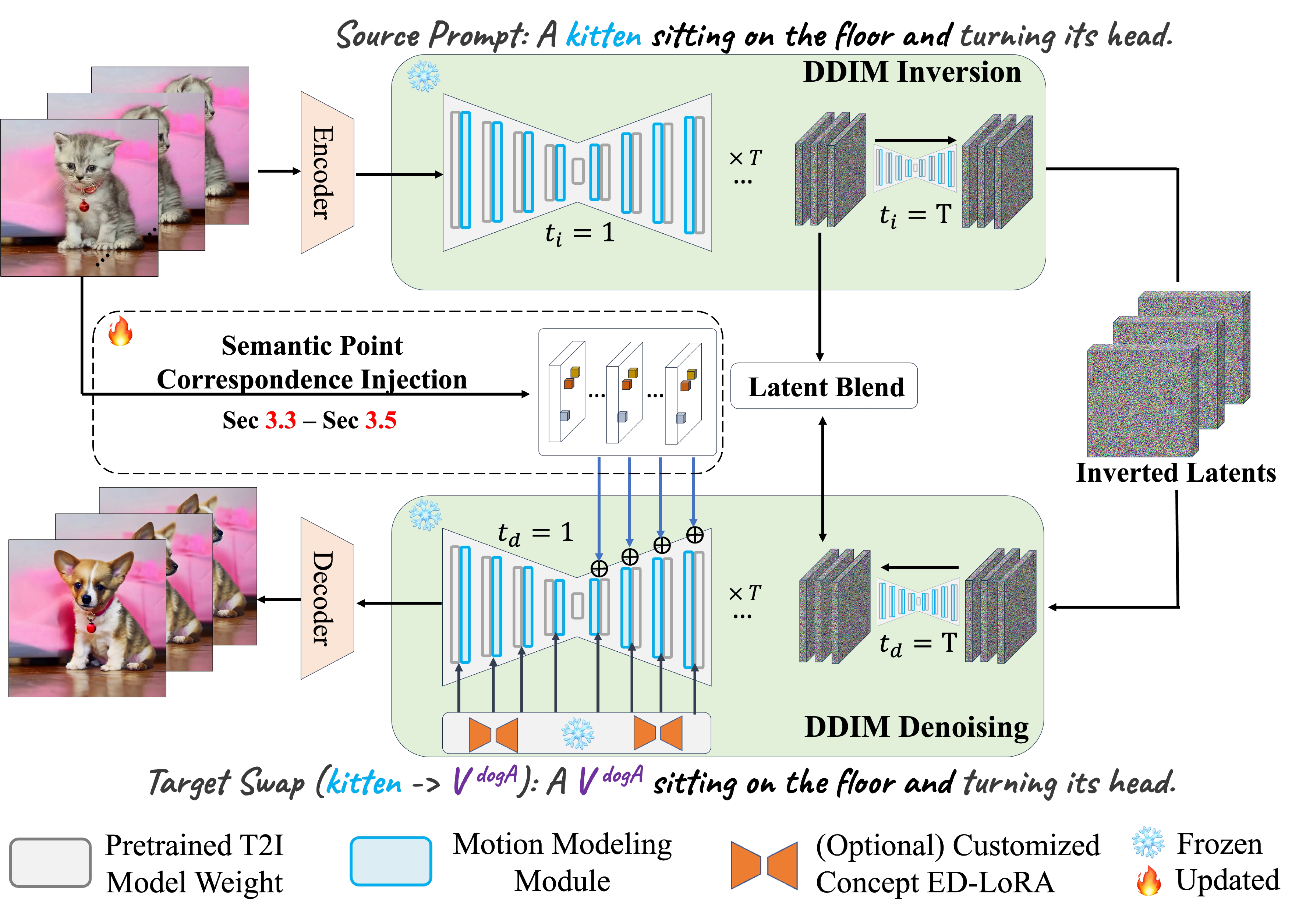}
    \vspace{-.12in}
    \caption{Overview of the VideoSwap pipeline for customized video subject swapping.}
\label{fig:pipeline_overview}
\vspace{-.1in}
\end{figure}

\begin{figure*}[!tb]
    \centering
\includegraphics[width=0.99\linewidth]{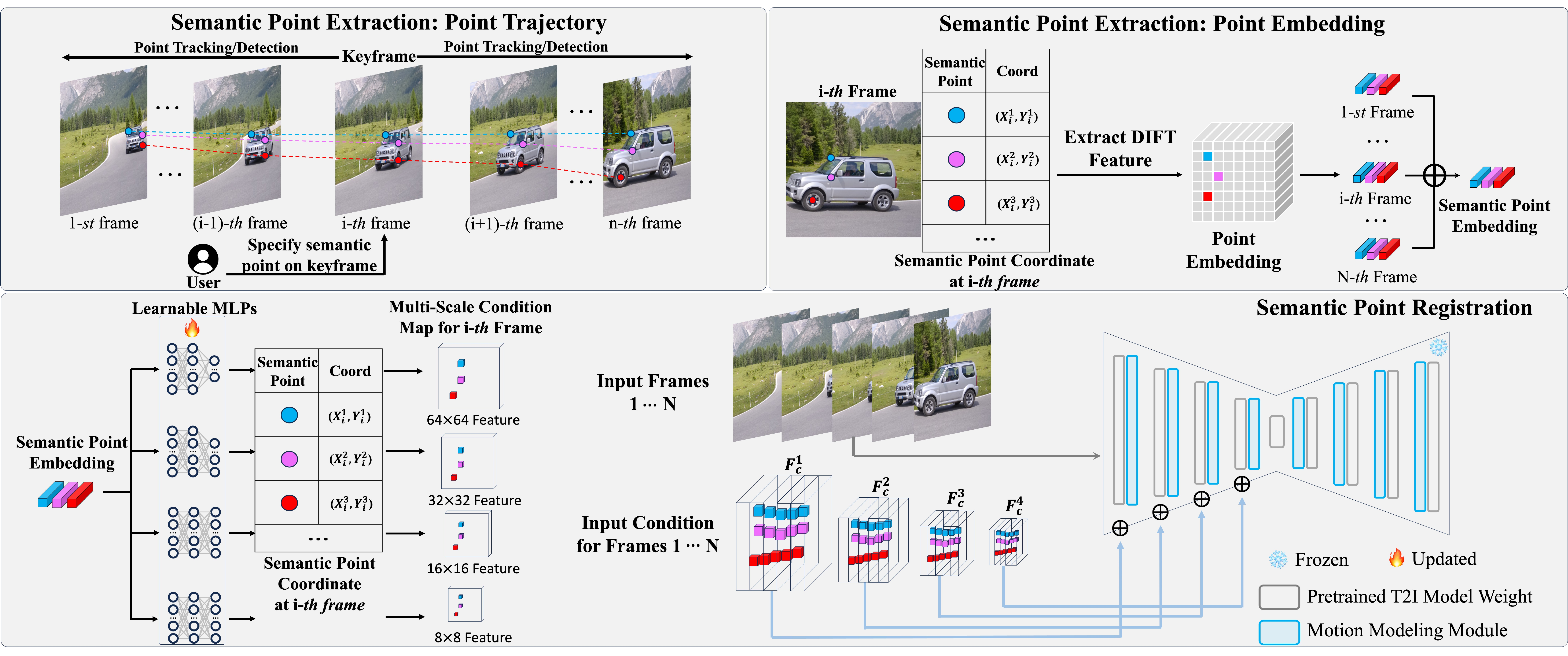}
    \vspace{-.16in}
    \caption{Pipelines for semantic point extraction (\secref{sec:source_point_extract}) and semantic point registration (\secref{sec:source_point_register}) in VideoSwap. In semantic point extraction, users define semantic points at a keyframe. We then extract the trajectory and embedding of those semantic points from the video. In semantic point registration, the semantic point embedding is projected by multiple 2-layer learnable MLPs, placed in empty features based on their coordinates, and then added element-wise to the diffusion model as motion guidance.}
    \label{fig:pipeline1}
    \vspace{-.1in}
\end{figure*}

\myPara{Observation 1:} \textbf{\textit{Semantic points optimized on source video frames have the potential to align the subject's motion trajectory and change the subject's shape.}}
As shown in \figref{fig:observation}(c, left), we drag the points on the cat's face. Despite this edited point map is not in the training data, the resulting image closely follows the adjusted semantic points, effectively generating the unseen pose of the subject. 
As shown in \figref{fig:observation}(c, right), when we drag the boundary semantic points of the car, the edited subject will also reshape to align with the semantic point.
This suggests the potential of utilizing semantic points to align the motion trajectory or change the shape when a sequence of dragged points for all video frames is accessible.

\myPara{Observation 2:} 
\textbf{\textit{Semantic points optimized on source video frames can transfer across semantic and low-level changes.}}
As demonstrated in \figref{fig:observation}(d), when we replace the source subject with different semantics or modify the low-level information in the prompts, the optimized semantic point can also control the pose or shape of the target concept. This suggests that the semantic point can be transferred across both semantic and low-level changes.

\subsection{Overview of VideoSwap}
\label{sec:overview}
\subsubsection{Task Formulation}

In this paper, we focus on customized video subject swapping, with the goal of \textbf{\textit{subject replacement}} and \textbf{\textit{background preservation}}.
Subject replacement requires preserving the identity of the target subject in the swapped results, encompassing both its appearance and shape. Simultaneously, background preservation requires the unedited background area to remain the same with the source video. The primary challenge of this task lies in aligning the motion trajectory of the source subject while preserving the identity of the target concept, particularly its shape.

\subsubsection{Overall Pipeline}
The VideoSwap pipeline is illustrated in \figref{fig:pipeline_overview}. Following the latent diffusion model~\cite{rombach2022high}, we encode the source video with a VAE encoder to obtain the latent space representation $z_0$. Subsequently, DDIM inversion~\cite{dhariwal2021diffusion, song2020denoising} is applied to transform the clean latent $z_0$ back to the noisy latent $z_T$. After obtaining the DDIM inverted noise $z_T$, we replace the source subject in the text prompt with the target subject and denoise it using the DDIM scheduler~\cite{song2020denoising}.
In this denoising process, we introduce semantic point correspondence to guide the subject's motion trajectory, as detailed in \secref{sec:source_point_extract}-\secref{sec:point_propagation}. To preserve the unedited background, we leverage the concept of latent blending~\cite{Avrahami_2022_CVPR, avrahami2023blendedlatent}, further explained in the \secref{sec:latent_blend}. Additionally, we incorporate the following designs:

\myPara{Adopting Motion Layers.} 
We integrate the motion layers~\cite{guo2023animatediff, blattmann2023align} into the image diffusion model to ensure essential temporal consistency for video editing.

\myPara{Supporting Predefined and Customized Concepts.} We support both predefined concepts from the pretrained model and customized concepts. 
To create customized concepts, we train ED-LoRA~\cite{gu2023mix} on a set of representative images to encode their identity. After training, these concept ED-LoRAs can be used at the inference time.

\subsection{Semantic Point Extraction}
\label{sec:source_point_extract}
We first extract the trajectories of semantic points and their associated semantic embeddings from the source video.

\begin{figure*}[!tb]
    \centering
\includegraphics[width=0.99\linewidth]{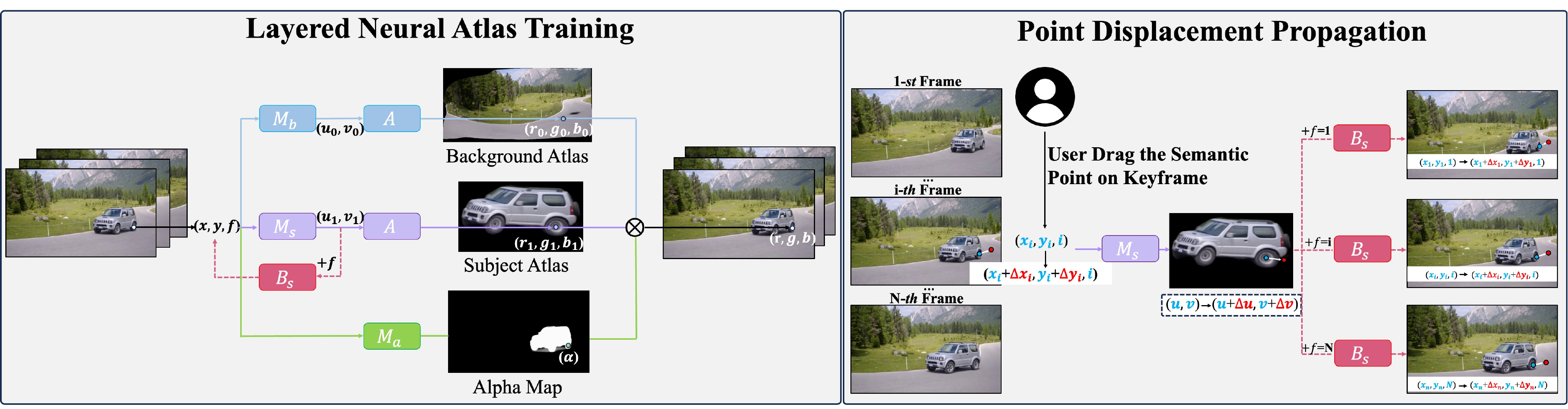}
    \vspace{-.14in}
    \caption{Point displacement propagation based on layered neural atlas (LNA)~\cite{kasten2021layered,huang2023inve}. Once a trained LNA is provided, users can drag a semantic point at the keyframe, and this displacement is consistently propagated to every frame through the canonical space of the LNA.}
    \vspace{-.1in}
\label{fig:pipeline_drag}
\end{figure*}

\myPara{Point Trajectory Extraction.} 
As depicted in \figref{fig:pipeline1}, for a video containing $N$ frames, users specify $K$ semantic points at a keyframe $i$. These user-defined semantic points are then propagated to the remaining $N-1$ frames using a point tracker~\cite{karaev2023cotracker} or detector~\cite{cao2017realtime}. Subsequently, the motion trajectory of all semantic points in the entire video is obtained and represented as $\mathbf{P}_{coord}=\{Tra(k)|k = 1...K\}$, where $Tra(k)=\{(x^k_n, y^k_n, n)|n=1...N\}$ represents the motion trajectory of semantic point $k$ across all $N$ frames.

\myPara{Point Embedding Extraction.}
To leverage semantic point correspondence, it is crucial to associate each point with its semantics. Specifically, we extract the DIFT~\cite{tang2023emergent} feature $\mathbf{D}_n$ for each frame $n$. Subsequently, the point embedding for semantic point $k$ at frame $n$ is obtained as $\mathbf{v}^k_n = \mathbf{D}_n(x^k_n, y^k_n)$, where $(x^k_n, y^k_n)$ is retrieved from $\mathbf{P}_{coord}$. Following this, we aggregate the point embeddings acquired from all $N$ frames to obtain the final embedding for each semantic point $k$ by $\mathbf{v}^k=\frac{1}{N}\sum_{n=1}^N \mathbf{v}^k_n$. Finally, we obtain the semantic embedding for all semantic points, succinctly represented as $\mathbf{P}_{emb} = \{\mathbf{v}^k | k = 1...K\}$.

\subsection{Semantic Point Registration}
\label{sec:source_point_register}

After acquiring the motion trajectory $\mathbf{P}_{coord}$ and the embedding $\mathbf{P}_{emb}$ for semantic points, we register these semantic points on the source video to enable them to offer motion guidance for the video subject.

\myPara{Sparse Motion Feature Creation.}
To utilize semantic points as guidance, we generate sparse motion features infused with semantic point embeddings, making them compatible with the Unet encoder.
We denote $\mathbf{F}_{enc}=\{\mathbf{F}_{enc}^1, \mathbf{F}_{enc}^2, \mathbf{F}_{enc}^3, \mathbf{F}_{enc}^4\}$ as the multi-scale intermediate feature of the Unet encoder.
For an input VAE-encoded latent with spatial-temporal size $(H,W,N)$, the feature size of $\mathbf{F}_{enc}^l$ for each Unet stage $l \in [1,4]$ can be computed as $(H/2^l, W/2^l, N)$.

As depicted in \figref{fig:pipeline1} and detailed in \algref{alg:motion_feature}, we create a series of multi-scale conditional features $\mathbf{F}_{c}=\{\mathbf{F}_{c}^1, \mathbf{F}_{c}^2, \mathbf{F}_{c}^3, \mathbf{F}_{c}^4\}$. Notably, $\mathbf{F}_{c}$ shares the same size as $\mathbf{F}_{enc}$ and is initialized with zero vectors.
And we introduce a series of learnable MLPs $\phi=\{\phi^l|l\in\{1,2,3,4\}\}$, each of $\phi^l$ project the point embedding to match the feature dimension of the corresponding $\mathbf{F}_c^l$.
Then, for each point $(x^k_n, y^k_n, n) \in \mathbf{P}_{coord}$, 
we compute its corresponding spatial-temporal coordinate $(x, y, n)$ at $l$-th Unet stage and assign the projected embedding based on the coordinate by $\mathbf{F}_c^l(x, y, n) = \phi^l(\mathbf{P}_{emb}(k))$.

It is crucial to emphasize that the motion feature $\mathbf{F}_c$ demonstrates high sparsity, with only the semantic point trajectories containing the feature embeddings. Finally, $\mathbf{F}_c$ are added element-wise into the intermediate feature $\mathbf{F}_{enc}$ of Unet encoder as motion guidance:
{\setlength{\abovedisplayskip}{1pt}
\setlength{\belowdisplayskip}{1pt}
\begin{equation}
    \mathbf{F}_{enc}^l = \mathbf{F}_{enc}^l + \mathbf{F}_{c}^l, l\in\{1,2,3,4\}.
\end{equation}}

{
\begin{algorithm}
  \caption{Sparse Motion Feature Creation}
\label{alg:motion_feature}
  \textbf{Input:}\\ 
     \quad 1. Point Trajectory $\mathbf{P}_{coord}$,\\
     \quad 2. Point Embedding $\mathbf{P}_{emb}$,\\
     \quad 3. Learnable MLPs $\phi=\{\phi^l|l\in\{1,2,3,4\}\}$ \\
  \textbf{Output:}\\
     \quad Motion Feature $\mathbf{F}_c=\{\mathbf{F}_c^l|l\in\{1,2,3,4\}\}$
  
  Initialize: $\mathbf{F}_c=\{\mathbf{F}_c^l=\mathbf{0}| l\in\{1,2,3,4\}\}$\;

  \For{$l \leftarrow 1$ \KwTo $4$}{
    \ForEach{$(x_n^k, y_n^k, n)$ in $\mathbf{P}_{coord}$}{
        $x, y = round(x_n^k / 2^l), round(y_n^k / 2^l)$\;
        $\mathbf{F}_c^l(x,y,n) = \phi^l(\mathbf{P}_{emb}(k))$
    }
  }
\end{algorithm}
}

\myPara{Semantic Point Registration on Source Video.}
Our objective is to optimize the projection MLPs ($\phi$) to facilitate better motion guidance from semantic points. This optimization objective is defined as
{\setlength{\abovedisplayskip}{1pt}
\setlength{\belowdisplayskip}{1pt}
\begin{equation}
    \footnotesize
    \min_\phi E_{ \epsilon\sim N(0, I), t\sim U(T_{min}, T)}||[\epsilon-\epsilon_\theta(z_t,t,p, \phi(\mathbf{P}_{emb}))]\odot \Omega(\mathbf{P}_{coord})||^2_2,
    \label{eq:objective}
\end{equation}}where $p$ represents the embedding for the text prompt, and $\Omega(\mathbf{P}_{coord})$ denotes the binary mask that only turns on around the semantic point.

\begin{figure*}[!tb]
    \centering
\animategraphics[width=0.99\linewidth,loop]{8}{imgs/comp_baseline/}{00001}{00016}
    \vspace{-.13in}
    \caption{Comparison of VideoSwap with several baselines built upon the same foundational model. The only difference lies in adopting different motion guidance.
We encourage readers to \textcolor{magenta}{click and play} the video clips in this figure using Adobe Acrobat.}
    \label{fig:baseline}
    \vspace{-.1in}
\end{figure*}

We adopt two techniques to enhance the learning of semantic point correspondences in Eq.~\eqref{eq:objective}. The first technique is \textbf{\textit{semantic-enhanced schedule}}, controlled by $T_{min}$. We set $T_{min}=T/2$ to enhance the learning at the higher timestep, which prevents overfitting to low-level details and improves semantic point alignment.
The second technique, \textbf{\textit{point patch loss}}, constrains the computation of the loss to a local patch near each semantic point, which reduces structure leakage into the target swap. This is implemented by the a loss mask $\Omega$ in Eq.~\eqref{eq:objective}. 

\subsection{User-Point Interaction at Inference Time}
\label{sec:point_propagation}
In \secref{sec:source_point_extract}, users define semantic points at the source video keyframe, and we subsequently extract their trajectory and semantic embedding in the video. To utilize these semantic points as correspondences, we register them on the source video in \secref{sec:source_point_register}. Following these two steps, these semantic points become applicable for controlling the motion of the target object. In this section, we introduce user-point interaction to address various semantic point correspondence.

\myPara{Adopting Source Point Sequence.} 
If there exist one-to-one semantic point correspondence between the source subject and the target subject, such as swapping the {\color{cyan}\textit{dog}} with {\color{Fuchsia}$V^{catA}$} as illustrated in \figref{fig:teaser1}, we directly use the source point sequence as the motion guidance.

\myPara{Removing Parts of Semantic Point.}
When there only exists partial semantic point correspondence between the source subject and the target subject, we can remove redundant points to loosen shape constraints. For example, in scenarios such as swapping an {\color{cyan}\textit{airplane}} for a {\color{Fuchsia}\textit{helicopter}}, as depicted in \figref{fig:teaser1}, we remove semantic points associated with the airplane's wings, given that helicopters typically lack wings. The remaining semantic points on the airplane's nose and tail are retained as motion guidance.

\myPara{Dragging of Semantic Point.}
In situations where there exists semantic point correspondence between the source and target subjects, but misalignment occurs due to shape morphing,  we provide users the option to manually drag the semantic points on a keyframe for better alignment of the shape changes. For instance, when swapping a {\color{cyan}\textit{jeep}} (tall and narrow) for a {\color{Fuchsia}$V_{carA}$} (sports car, low and wide) as illustrated in \figref{fig:teaser1}, users can drag the semantic points to accurately reflect the shape change.

Editing the shape change by dragging semantic points on a single frame is straightforward. However, consistently propagating these semantic point displacement over time is non-trivial, mainly because of the complex camera and subject motion in the video. Therefore, we introduce point displacement propagation to solve this problem.

As illustrated in \figref{fig:pipeline_drag}, we follow the Layered Neural Atlas (LNA)~\cite{kasten2021layered,huang2023inve} to learn the canonical space, as detailed in the \secref{sec:nla_training}. 
Following LNA~\cite{kasten2021layered,huang2023inve}, we establish a forward coordinate mapping from the video to the canonical space, denoted as $M$: $(x, y, f) \rightarrow (u, v)$, along with its corresponding backward mapping $B$: $(u, v, f) \rightarrow (x, y)$.

Given a semantic point, with coordinates at the keyframe $f_{key}$ represented as $(x, y, f_{key})$, its trajectory over time can be expressed as a function of time $f$: $(x(f), y(f)) = P(f)$.
Suppose a user drag it to a new position at $(x + dx, y + dy, f_{key})$,
we aim to estimate the edited trajectory $P'(f)$ for $f = \{0, ..., N\}$. 
We resort to LNA's representation, and first compute a linearized estimation of its shifted position on the canonical coordinate:
{\setlength{\abovedisplayskip}{1pt}
\setlength{\belowdisplayskip}{1pt}
\begin{equation}
    [du, dv]^T = J_M(x, y, f_{key}) [dx, dy]^T,
\end{equation}}where $J_{M}$ denote the Jacobian matrix with respect to $(x, y)$.
Next, at a given time $f$, we estimate the edited coordinate in the pixel space as 
{\setlength{\abovedisplayskip}{1pt}
\setlength{\belowdisplayskip}{1pt}
\begin{equation}
P'(f) ~= P(f) + J_B(u, v, f)[du, dv]^T,
\end{equation}}where $(u, v)=B(x,y,f_{key})$ and $J_{B}$ denote the Jacobian matrix with respect to $(u, v)$. 
In practice, we approximate the Jacobian computation by
\begin{align}
J_M &= \begin{bmatrix}
M_s(x+\varepsilon
, y, f) - M_s(x, y, f)\\
M_s(x, y+\varepsilon
, f) - M_s(x, y, f) 
\end{bmatrix}^T
\begin{bmatrix}
1 / \varepsilon
\\
1/ \varepsilon
\end{bmatrix},\\
J_B &= \begin{bmatrix}
B_s(u+\varepsilon
, v, f) - B_s(u, v, f)\\
B_s(u, v+\varepsilon
, f) - B_s(u, v, f) 
\end{bmatrix}^T
\begin{bmatrix}
1 / \varepsilon
\\
1/ \varepsilon
\end{bmatrix},
\end{align}where $\varepsilon$ represents the small coordinate shift.
We then use this edited trajectory $P'(f)$ for the dragged semantic point during inference.

\begin{table}[!tb]
\centering

\resizebox{\linewidth}{!}{\begin{tabular}{cl|cccc}
\toprule
\multicolumn{2}{l|}{\textbf{Methods/Metrics}} & \multicolumn{1}{c}{\begin{tabular}[c]{@{}c@{}}\textbf{Subject} \\ \textbf{Identity}\end{tabular}} & \multicolumn{1}{c}{\begin{tabular}[c]{@{}c@{}}\textbf{Motion} \\ \textbf{Alignment}\end{tabular}} & \multicolumn{1}{c}{\begin{tabular}[c]{@{}c@{}}\textbf{Temporal}\\ \textbf{Consistency}\end{tabular}} & \multicolumn{1}{c}{\begin{tabular}[c]{@{}c@{}}\textbf{Overall}\\ \textbf{Preference}\end{tabular}} \\ \hline
&\multicolumn{4}{|c}{\quad\quad\quad\quad\quad\quad\quad\textbf{Compare to Previous Video Editing Methods}}\\\cline{2-6}
\multicolumn{1}{c|}{\multirow{7}{*}{\begin{tabular}[c]{@{}c@{}}\rotatebox{90}{\textbf{VideoSwap} v.s. }\end{tabular}}} & Tune-A-Video & \textbf{80\%} v.s. 20\% & \textbf{72\%} v.s. 28\% & \textbf{80\%} v.s. 20\% & \textbf{78\%} v.s. 22\%\\ 
\multicolumn{1}{c|}{} & FateZero & \textbf{74\%} v.s. 26\% & \textbf{67\%} v.s. 33\% & \textbf{73\%} v.s. 27\% & \textbf{70\%} v.s. 30\%\\
\multicolumn{1}{c|}{} & Text2Video-Zero & \textbf{76\%} v.s. 24\% & \textbf{71\%} v.s. 29\% & \textbf{80\%} v.s. 20\% & \textbf{80\%} v.s. 20\%\\
\multicolumn{1}{c|}{} & Rerender-A-Video & \textbf{81\%} v.s. 19\% & \textbf{72\%} v.s. 28\% & \textbf{84\%} v.s. 16\% & \textbf{84\%} v.s. 16\%\\ \cline{2-6}
&\multicolumn{4}{|c}{\quad\quad\quad\quad\quad\quad\quad\textbf{Compare to Baselines on AnimateDiff}}\\\cline{2-6}
\multicolumn{1}{c|}{} & w/ DDIM & \textbf{70\%} v.s. 30\% & \textbf{74\%} v.s. 26\% & \textbf{69\%} v.s. 31\% & \textbf{73\%} v.s. 27\%\\ 
\multicolumn{1}{c|}{} & w/ DDIM+TAV & \textbf{68\%} v.s. 32\% & \textbf{60\%} v.s. 40\% & \textbf{66\%} v.s. 34\% & \textbf{69\%} v.s. 31\%\\ 
\multicolumn{1}{c|}{} &  w/ DDIM+T2I-Adapter & \textbf{66\%} v.s. 34\% & \textbf{59\%} v.s. 41\% & \textbf{65\%} v.s. 35\% & \textbf{66\%} v.s. 34\%\\ \bottomrule
\end{tabular}}
\vspace{-.1in}
\caption{Human Evaluation on Video Subject Swapping Results.}
\vspace{-.25in}
\label{tab:human_eval}
\end{table}

\begin{figure*}[!tb]
    \centering
\animategraphics[width=.99\linewidth,loop]{8}{imgs/ablation_study/}{00001}{00016}
    \vspace{-.16in}
    \caption{Ablation Study of our VideoSwap.
    We encourage readers to \textcolor{magenta}{click and play} the video clips in this figure using Adobe Acrobat.}
    \label{fig:ablation}
    \vspace{-.1in}
\end{figure*}

\section{Experiments}

We implement our method using the Latent Diffusion Model~\cite{rombach2022high} and adopt the motion layer in AnimateDiff~\cite{guo2023animatediff} as the foundational model. All videos consist of 16 frames. The primary time cost is registering semantic points in the video, which requires about 3 minutes per video. Additional implementation details, as well as time and memory cost analyses, are included in the \secref{sec:time_cost}.

\subsection{Qualitative Comparison}

\myPara{Comparison with the State-of-the-Art.}
We qualitatively compare to Tune-A-Video~\cite{wu2023tune}, FateZero~\cite{qi2023fatezero}, Rerender-A-Video~\cite{yang2023rerender}, TokenFlow~\cite{geyer2023tokenflow} and StableVideo~\cite{chai2023stablevideo} in \figref{fig:teaser1}. Previous methods are less effective in revealing the correct shape of the target subject. Compared to them, VideoSwap can achieve a significant shape change while aligning the source motion trajectory.

\myPara{Comparison with Baselines on AnimateDiff.}
As most state-of-the-art methods are based on image diffusion models, we also compare VideoSwap to several baselines on the AnimateDiff. The only distinctions from VideoSwap lie in different motion guidance, as shown in the results in \figref{fig:baseline}:
\begin{itemize}
    \item DDIM: The DDIM sampling without other motion guidance cannot produce the correct motion trajectory.
    \item DDIM+Tune-A-Video: If we tune the model as~\cite{wu2023tune} to inject source motion, it achieves correct motion but suffers from severe structure and appearance leakage.
    \item DDIM+T2I-Adapter: If we add spatial controls~\cite{mou2023t2i}, such as depth, to control the editing, we observe that 1) the shape is restricted by the source, and 2) the deformable motion cannot follow the source video.
\end{itemize}
Compared to all constructed baselines, our VideoSwap with semantic point correspondence can effectively align the motion trajectory while preserving the target concept's identity.

\subsection{Quantitative Comparison}
We conduct both automatic and human evaluations to quantitatively compare VideoSwap with previous state-of-the-art methods and several baselines on AnimateDiff. Detailed evaluation settings and automatic evaluation results are provided in the \secref{sec:supp_quantitative}. For human evaluation, we distribute 1000 questionnaires on Amazon Mturk to assess various criteria in customized video subject swapping.
From the human evaluation results in \tabref{tab:human_eval}, our method achieves a clear advantage over the compared methods.

\subsection{Ablation Study}

\myPara{Sparse Motion Feature.}
There are several variants for encoding semantic points to generate motion guidance. The most straightforward approach is to encode the point map of semantic points using T2I-Adapter~\cite{mou2023t2i}. However, this method leads to severe overfitting, as illustrated in \figref{fig:ablation}(a). The issue arises because the encoded feature added to the diffusion model is non-sparse, and the background also contains features that may overfit to the source video.
We use MLPs to encode point embeddings, positioning them in the empty feature to create sparse motion features for guidance without compromising video quality. 
Regarding point embeddings, we opt for DIFT embeddings, which inherently carry robust semantics. Compared to randomly initialized learnable embeddings, our approach achieves similar results with 3$\times$ less time for point registration step.

\myPara{Point Patch Loss.}
In the semantic point registration, we employ a point patch loss to reconstruct the local patch surrounding each semantic point. If we omit the point patch loss and opt to directly reconstruct the entire video, we observe that the source structure leaks into the target swapped results and thus produce artifacts, as depicted in \figref{fig:ablation}(b).

\myPara{Semantic-Enhanced Schedule.}
Our goal is to employ semantic points to transfer motion trajectories, acting as a linkage between the source and target subjects. Therefore, we aim for the semantic points to emphasize high-level semantic alignment without transferring low-level details. This objective is achieved by registering points only during the early sampling steps, \ie, $T_{\text{min}}=\frac{T}{2}$ in Eq.~\eqref{eq:objective}. As shown in \figref{fig:ablation}(c), this technique prevents the model from learning excessive low-level details and enhances semantic point alignment. 

\myPara{Drag-based Point Control.}
As illustrated in \figref{fig:ablation}(d), our goal is to swap the {\color{cyan}\textit{black swan}} to the {\color{ForestGreen}\textit{duck}}. If we directly use the source point sequence as guidance, the {\color{ForestGreen}\textit{duck}}'s neck conforms to the shape of the {\color{cyan}\textit{black swan}}, resulting in an inferior identity. However, by employing the proposed point displacement propagation, we can drag the semantic point at the keyframe, ensuring a consistent motion trajectory after dragging. Utilizing the dragged semantic point trajectory as motion guidance allows us to accurately establish the identity of the {\color{ForestGreen}\textit{duck}}.

\section{Conclusion}

This paper uncovers the potential of semantic point correspondence in aligning motion trajectories and altering the subject's identity in video editing. From there, we present VideoSwap, a framework that minimizes human intervention while utilizing semantic point correspondences for customized video subject swapping. Through user-point interactions like point removal or dragging, we address various semantic point correspondence. VideoSwap facilitates shape changes in the target swap while aligning the motion trajectory with the source subject, demonstrating state-of-the-art results in customized video subject swapping.

{
    \small
    \bibliographystyle{ieeenat_fullname}
    \bibliography{main}
}

\clearpage
\setcounter{page}{1}
\maketitlesupplementary

\section{Additional Details about Methods}

\subsection{Latent Blend}
\label{sec:latent_blend}
Given our focus on subject swapping, where the objective is to maintain the unedited background region identical to the source video, this is achieved through latent blend~\cite{guo2023animatediff, blattmann2023align}, as shown in \figref{fig:pipeline_overview}.

The key idea is that the latent noise in DDIM denoising and DDIM inversion provides information for the swapped subject and background, respectively. These two latent noises can be blended using a mask that indicates the foreground region, thus blending the swapped target with the source background.

To initiate the process, we acquire the foreground mask for timestep $t$ as $\mathcal{M}^t = \mathcal{M}^t_i \cup \mathcal{M}^t_d$, formed by merging the subject masks $\mathcal{M}_i$ during inversion and $\mathcal{M}_d$ during denoising at the same timestep $t$. This subject mask is automatically generated through the cross-attention of the concept token, following the approach of Prompt2Prompt~\cite{hertz2022prompt}.

Subsequently, the foreground mask is used to blend the latent features, resulting in $z^t = (1-\mathcal{M}^t) \cdot z^t_i + \mathcal{M}^t \cdot z^t_d$, where $z^t_i$ and $z^t_d$ represent the latent features of timestep $t$ in DDIM inversion and DDIM denoising, respectively. Through latent blend, we can effectively preserve the unedited background in the source video.

\subsection{Layered Neural Atlas Training}
\label{sec:nla_training}
As mentioned in \secref{sec:point_propagation} of the main paper, we introduce interactive dragging on the key frame for handling point correspondence with shape morphing in customized video subject swapping. This function is supported by the learned canonical space of Layered Neural Atlas~\cite{kasten2021layered} (LNA). Here, we present a detailed formulation of LNA.

LNA~\cite{kasten2021layered} represents a video through the following three sets of parameterized MLPs:
\begin{enumerate}
    \item \textbf{Coordinate Mapping MLPs}. The coordinate mapping MLPs map the spatial-temporal coordinates of video pixels to the 2D canonical space (\ie, the UV map), denoted as $M$: $(x, y, f) \rightarrow (u, v)$. We employ separate mappings, $M_s$ and $M_b$, for the foreground subject and background, respectively. Additionally, following the approach of INVE~\cite{huang2023inve}, we include a background mapping $B_s$: $(u, v) \rightarrow (x, y, f)$ to learn the coordinate mapping of the foreground subject from the canonical space back to the video pixel.
    \item \textbf{Atlas MLPs}. The atlas MLPs, denoted as $A$: $(u, v) \rightarrow (r, g, b)$, learn to predict the color of the coordinates on the UV map.
    \item \textbf{Alpha MLPs}. 
    The alpha MLPs, denoted as $M_\alpha$: $(x, y, f) \rightarrow \alpha$, predict the blending ratio $\alpha$ of the color value from the subject atlas and background atlas.
\end{enumerate}
Based on these sets of learnable MLPs, the training objective of LNA is to reconstruct the RGB values of the source video, accompanied by the following regularization losses:
\begin{enumerate}
    \item \textbf{Rigidity Loss}. The rigidity loss encourages the learned mapping from pixel coordinates in the video to the 2D canonical space to exhibit local rigidity.
    \item \textbf{Consistency Loss}. The consistency loss encourages the mapping of corresponding video pixels across consecutive frames to be consistent, with correspondence estimated through pre-computed optical flow.
    \item \textbf{Sparsity Loss}. The sparsity loss encourages the many-to-one mapping from the video coordinates to the canonical coordinates, penalizing duplicate contents in the canonical space.

\end{enumerate}
We refer the reader to the LNA~\cite{kasten2021layered} paper for the complete formulation.

\subsection{Discussion the Relation to Human Keypoint}
The ControlNet~\cite{zhang2023adding} and T2I-Adapter~\cite{mou2023t2i} also incorporate control over human keypoints. These human keypoints can be viewed as a type of sparse semantic points, where the semantic position and total number of human keypoints are predefined by the existing pose detectors, and their semantic embedding for controlling the diffusion model is implicitly aligned through large-scale paired data. However, defining keypoints or collecting paired data for open-set concepts proves challenging due to the variability in semantic points. Therefore, our method provides a more generic framework for point-based video editing, with human keypoints serving as a specific use case within our framework.

\section{Experimental Details}
\subsection{Implementation Details}

We implement our method using the Latent Diffusion Model~\cite{rombach2022high} and incorporate the pretrained motion layer from AnimateDiff~\cite{guo2023animatediff} as the foundational model. All experiments are conducted on an Nvidia A100 (40GB) GPU.
All video samples consist of 16 frames with a time stride of 4, matching the temporal window of the motion layer in AnimateDiff. We crop the videos to two alternate resolutions ($H\times W$): $512\times 512$ and $448 \times 768$. For all experiments, we employ the Adam optimizer with a learning rate of 5e-5, optimizing for 100 iterations. Regarding the point patch loss, we use a patch size of $4\times 4$ around the semantic point.

\subsection{Time Cost Analysis}
\label{sec:time_cost}
In this section, we analyze the time cost of editing a video in VideoSwap. All time costs are calculated on an Nvidia A100 GPU to process a 16 frame video clip. 

\myPara{Time Cost of Preprocess.}
The preprocessing step involves (1) extracting point trajectories and their DIFT embeddings, and (2) registering those semantic points to guide the diffusion model, and (3) generate DDIM-inverted noise. The extraction of trajectories and embeddings takes approximately 30 seconds. The registration step requires 100 iterations, taking about 3 minutes. And the DDIM inversion of 50 steps takes approximately 30 seconds. To summarize, it takes about 4 minutes to pre-process a video for editing.

\myPara{Time Cost of Each Edit.}
Then for each edit, the time cost of VideoSwap remain the similar to AnimateDiff~\cite{guo2023animatediff}, necessitating 50 seconds with the latent blend technique. The introduction of semantic point correspondence does not notably increase the time cost, given its lightweight computation.

\myPara{Time Cost of User-Point Interaction.}
The time cost for user-point interaction (\eg, removing or dragging a point) can be negligible. Dragging a point at the keyframe only takes 1 seconds to propagate to all other frames through a learned layered neural altas (LNA).

\myPara{Extra Time Cost in Training LNA.}
Our support for drag-based editing is built upon a learned LNA of the given video. In contrast to the original LNA, which necessitates approximately 10 hours of training, we do not require full training as we only adopt the forward/backward coordinate mapping. This training process takes about 2 hours for a video.

\subsection{Memory Cost Analysis}
The overall memory cost is similar to AnimateDiff, where we don't incur significant additional memory costs, as our semantic points and MLPs are lightweight. It only requires a memory cost of 16/12 GB for point registration and inference, respectively.

\begin{table}[!tb]
\centering
\resizebox{\linewidth}{!}{\begin{tabular}{l|ccc}
\hline
\multirow{2}{*}{Methods/Metrics}
& \multicolumn{1}{c}{\multirow{2}{*}{\begin{tabular}[c]{@{}c@{}}Text\\ Alignment $(\uparrow)$\end{tabular}}} & \multicolumn{1}{c}{\multirow{2}{*}{\begin{tabular}[c]{@{}c@{}}Image\\ Alignment $(\uparrow)$\end{tabular}}} & \multicolumn{1}{c}{\multirow{2}{*}{\begin{tabular}[c]{@{}c@{}}Temporal \\ Consistency $(\uparrow)$\end{tabular}}} \\ 
& \multicolumn{1}{c}{} & \multicolumn{1}{c}{} & \multicolumn{1}{c}{} \\\hline
\multicolumn{4}{c}{\textbf{Compare to Previous Video Editing Methods}}\\
\hline
Tune-A-Video ~\cite{wu2023tune} & 25.34 & - & 95.79 \\
FateZero ~\cite{qi2023fatezero} & 
24.39 & - & 95.49 \\
Text2Video-Zero ~\cite{khachatryan2023text2video} & 24.85 & - & 95.02 \\
Rerender-A-Video ~\cite{yang2023rerender} & 24.99 & - & 92.28 \\
VideoSwap (Ours) & \textbf{26.87} & - & \textbf{95.93} \\
\hline
\multicolumn{4}{c}{\textbf{Compare to Baselines on AnimateDiff}}\\\cline{1-4}
w/ DDIM & 27.36 & 79.79 & 95.89 \\ 
w/ DDIM + TAV & 24.75 & 75.93 & 95.49 \\ 
w/ DDIM + T2I-Adapter & 25.86 & 77.54 & 95.50 \\
VideoSwap (Ours) & 26.87 & \textbf{79.87} & \textbf{95.93} \\ \bottomrule
\end{tabular}}
\vspace{-.15in}
\caption{Automatic Quantitative Evaluation on Video Subject Swapping Results.}
\label{tab:auto_eval}
\end{table}

\begin{figure*}[!tb]
    \centering
\includegraphics[width=0.99\linewidth]{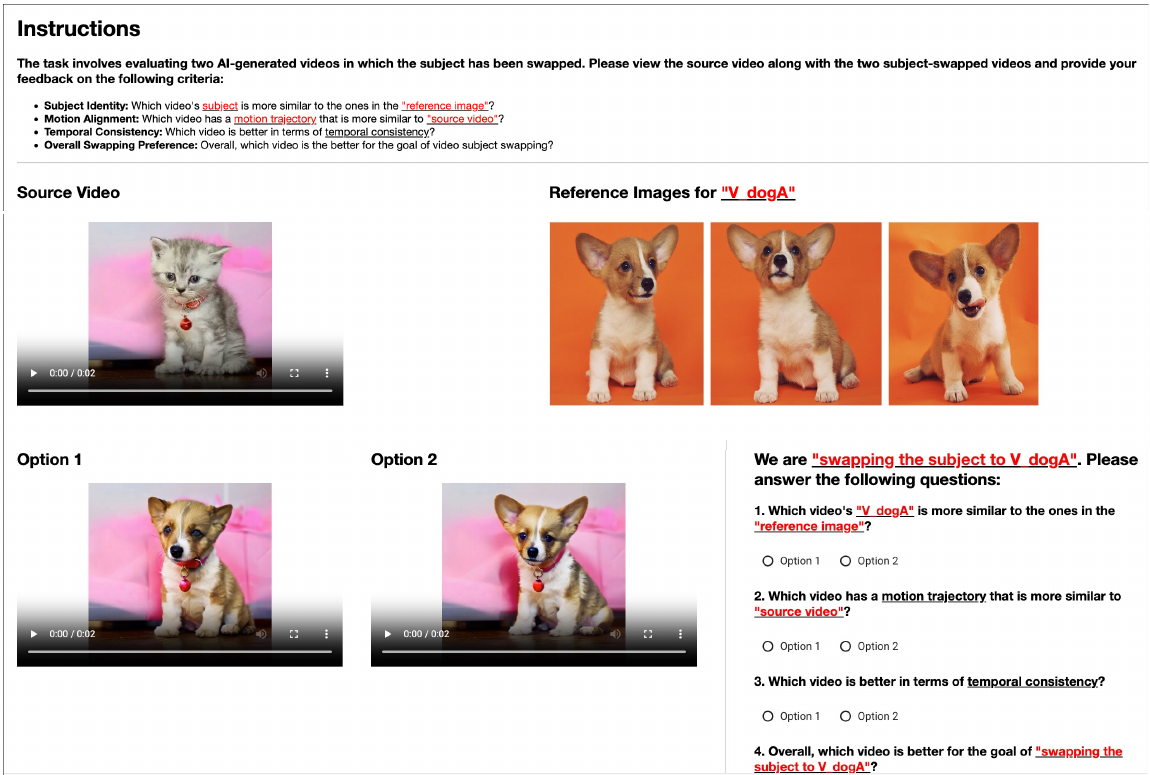}
    \vspace{-.1in}
    \caption{Human evaluation interface on Amazon Mturk. We provide the source video and reference images for target concept and ask user to select favorable video in terms of different criteria of video subject swapping.}
    \label{fig:interface}
\end{figure*}

\begin{table*}[!tb]
\centering

\resizebox{0.8\linewidth}{!}{\begin{tabular}{cl|cccc}
\toprule
\multicolumn{2}{l|}{\textbf{Methods/Metrics}} & \multicolumn{1}{c}{\begin{tabular}[c]{@{}c@{}}\textbf{Subject} \\ \textbf{Identity}\end{tabular}} & \multicolumn{1}{c}{\begin{tabular}[c]{@{}c@{}}\textbf{Motion} \\ \textbf{Alignment}\end{tabular}} & \multicolumn{1}{c}{\begin{tabular}[c]{@{}c@{}}\textbf{Temporal}\\ \textbf{Consistency}\end{tabular}} & \multicolumn{1}{c}{\begin{tabular}[c]{@{}c@{}}\textbf{Overall}\\ \textbf{Preference}\end{tabular}} \\ \hline
&\multicolumn{4}{|c}{\quad\quad\quad\quad\quad\quad\quad\textbf{Ablation of Sparse Motion Feature}}\\\cline{2-6}
\multicolumn{1}{c|}{\multirow{6}{*}{\begin{tabular}[c]{@{}c@{}}\rotatebox{90}{\textbf{VideoSwap} v.s. }\end{tabular}}} & Point Map + T2I-Adapter (100 iters) & \textbf{87\%} v.s. 13\% & \textbf{90\%} v.s. 10\% & \textbf{87\%} v.s. 13\% & \textbf{90\%} v.s. 10\% \\ 
\multicolumn{1}{c|}{} & Learnable Embedding + MLP (100 iters) & \textbf{87\%} v.s. 13\% & \textbf{95\%}  v.s. 5\% & \textbf{90\%}  v.s. 10\% & \textbf{90\%}  v.s. 10\%\\
\multicolumn{1}{c|}{} & Learnable Embedding + MLP (300 iters) & \textbf{52\%} v.s. 48\% & \textbf{52\%} v.s. 48\% & \textbf{52\%} v.s. 48\% & \textbf{55\%} v.s. 45\% \\\cline{2-6}
&\multicolumn{4}{|c}{\quad\quad\quad\quad\quad\quad\quad\textbf{Ablation of Point Patch Loss}}\\\cline{2-6}
\multicolumn{1}{c|}{} & w/o. Point Patch Loss & \textbf{73\%} v.s. 27\% & \textbf{73\%} v.s. 27\% & \textbf{78\%} v.s. 22\% & \textbf{78\%} v.s. 22\% \\ \cline{2-6}
&\multicolumn{4}{|c}{\quad\quad\quad\quad\quad\quad\quad\textbf{Ablation of Semantic-Enhanced Schedule}}\\\cline{2-6}
\multicolumn{1}{c|}{} & w/o. Semantic-Enhanced Schedule & \textbf{85\%} v.s. 15\% & \textbf{90\%} v.s. 10\% & \textbf{90\%} v.s. 10\% & \textbf{87\%} v.s. 13\% \\ 
\bottomrule
\end{tabular}}
\vspace{-.14in}
\caption{Human Evaluation for Ablation Study in VideoSwap. VideoSwap utilizes DIFT embedding + MLP (100 iterations) and incorporates the point patch loss and a semantic-enhanced schedule to improve the learning of semantic point correspondence.}
\label{tab:human_eval_ablation}
\end{table*}

\section{Quantitative Evaluation}
\label{sec:supp_quantitative}
\subsection{Dataset and Evaluation Setting}
We collect 30 videos from Shutterstock and DAVIS~\cite{perazzi2016benchmark}. Each category—human, animal, and object—comprises 10 videos. Besides, we gather 13 customized concepts: 5 for human characters, 3 for animals, and 5 for objects. Due to legal concerns, we cannot demonstrate qualitative results involving human characters. For each source video, we adopt 8 predefined concepts and 2-5 customized concepts as swap targets, yielding approximately 300 edited results.
For comparison to previous video-editing methods that don't support customized concepts, we only compute the metric on predefined concepts. In comparison to the baselines built upon AnimateDiff~\cite{guo2023animatediff}, we compute the metric on both predefined concepts and customized concepts.

\subsection{Automatic Evaluation by CLIP-Score}
We conduct a quantitative evaluation using the automatic metric, CLIP-Score~\cite{hessel2021clipscore}. The metric includes text alignment and temporal consistency, following~\cite{wu2023cvpr}. Additionally, for customized concepts, we follow Custom Diffusion~\cite{kumari2022multi} to compute pairwise image alignment between each edited frame and each reference concept image.
The results are summarized in \tabref{tab:auto_eval}. In comparison to previous video editing methods, VideoSwap demonstrates the best text alignment and temporal consistency. Moreover, when compared to baselines built on AnimateDiff, we achieve superior image alignment and temporal consistency. However, it is important to note that CLIP-Score is primarily based on frame-wise computation and may not align well with human perception, as discussed in EvalCrafter~\cite{liu2023evalcrafter}. Therefore, we present these results for reference purposes and primarily evaluate and compare using human evaluation.

\begin{figure*}[!tb]
    \centering
\includegraphics[width=0.99\linewidth]{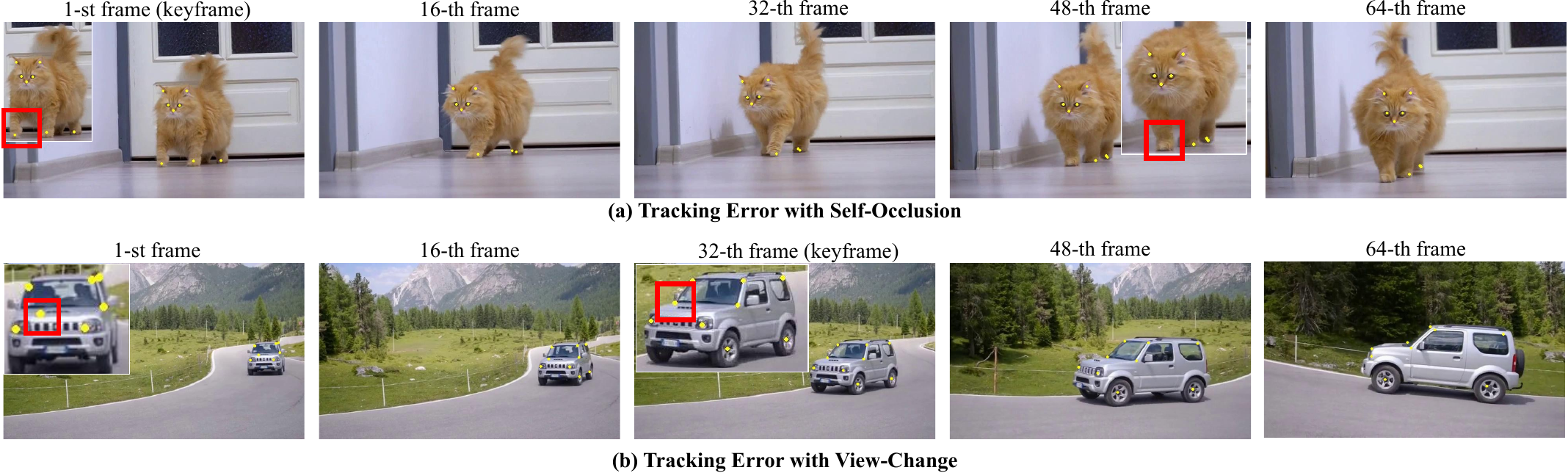}
    \vspace{-.1in}
    \caption{
Limitations in point tracking inherited from Co-Tracker~\cite{karaev2023cotracker} in scenarios involving self-occlusion and significant view changes.}
    \label{fig:limitation}
\end{figure*}

\subsection{Human Evaluation Interface}
We primarily conduct human evaluations to compare different methods based on several criteria: subject identity, motion alignment, temporal consistency, and overall swapping preference.
As depicted in ~\figref{fig:interface}, we present the source video and reference images for the target concept in the interface and ask users to select their preferred video based on various criteria related to customized video subject swapping.
We distribute 1000 questionnaires on Amazon Mturk. The human evaluation results in Table. \textcolor{red}{1} of the main paper clearly demonstrate our advantage.

\subsection{Human Evaluation for Ablation Study}
We employ human evaluation to quantitatively assess various variants of our methods, and the results are summarized in \tabref{tab:human_eval_ablation}. In terms of creating sparse motion features, our DIFT embedding significantly outperforms point map + T2I-Adapter and the learnable embedding + MLP with the same registration iterations. In comparison to the learnable embedding and MLP, our explicit DIFT embedding already contains sufficient semantic information, requiring 3$\times$ less time to achieve similar preference.
The introduction of the point patch loss and semantic-enhanced schedule further enhances VideoSwap, leading to higher preferences compared to variants without these enhancements.

\section{Qualitative Evaluation}

All our qualitative results and analysis are presented in the \url{https://videoswap.github.io/supplementary/videoswap_supp.html}. Please visit this webpage for more details.

\section{Limitation and Future Works}
\subsection{Limitation Analysis}
The limitation of VideoSwap is inherited from inaccurate point tracking and an imperfect canonical space representation of Layered Neural Atlas.

\myPara{Inaccurate Point Tracking by Co-Tracker.}
VideoSwap relies on accurate point trajectory extraction. However, the existing point tracking method Co-Tracker~\cite{karaev2023cotracker} is not stable enough when the video contains self-occlusion and large view changes, as shown in \figref{fig:limitation}(a) and \figref{fig:limitation}(b).
To address this issue, users may choose to remove inaccurate semantic points; however, this would result in less motion alignment. Nevertheless, since tracking any point is a newly formed problem, any progress in this area can seamlessly integrate into VideoSwap.

\myPara{Imperfect Canonical Space by Layer Neural Atlas.}
As discussed in Layered Neural Atlas (LNA)~\cite{kasten2021layered}, LNA fails to represent videos involving 3D rotations and non-rigid motion with self-occlusion. VideoSwap resorts to LNA to propagate the dragged point displacement. Therefore, due to the limitations of LNA, we cannot support drag-based interaction in such cases. Improvement in LNA representation will further broaden support for drag-based video editing.

\myPara{Time Cost for Interactive Editing.}
The time cost of VideoSwap prohibits its use for real-time interactive editing. Setting up semantic points for a video takes approximately 4 minutes. And to support drag-based editing, an additional 2 hours are required to prepare the LNA for the given video. Furthermore, constrained by diffusion model sampling, it takes about 50 seconds to perform an edit, falling short of real-time editing. We anticipate that advancements in neural field acceleration~\cite{huang2023inve,muller2022instant,kerbl20233d} and diffusion model distillation~\cite{luo2023latent, song2023consistency, salimans2022progressive} will significantly reduce the preprocess cost and enhance speed for real-time interactive editing.

\subsection{Future Works}

VideoSwap embarks on video editing with shape change. With semantic points as correspondence, VideoSwap can support interactive editing for large shape changes while aligning motion trajectories. We list several promising directions motivated by VideoSwap.

\myPara{Interactive Video Editing.}
VideoSwap supports drag-based interaction at the keyframe, propagating the dragged displacement throughout the entire video and obtaining the source and dragged trajectories with similar motion.
As we can obtain the source point trajectory and target point trajectory, future work may extend the idea of DragGAN~\cite{pan2023drag} to the video domain for drag-based real video editing.

\myPara{Video Editing with Shape Change.}
VideoSwap has demonstrated promising results in swapping the subject in the source video with a target concept that may have a different shape. In our paper, we focus on the swapping foreground subject, without considering background swapping or stylization. Further research could delve into a more general framework for video editing involving shape changes, thereby enhancing the flexibility of the video editing.

\myPara{Application Based on Customized Video Editing.}
VideoSwap has shown promising results in swapping the subject in the source video with a target concept with customized identity. Future work may further investigate its application in movie generation and storytelling by fixing subjects' identities.

\subsection{Potential Negative Social Impact}
This project aims to provide the community with an effective method to swap their customized concept into the video.
However, a risk exists wherein malicious entities could exploit this framework to create deceptive video with real-world figures, potentially misleading the public. This concern is not unique to our approach but rather a shared consideration in other concept customization methodologies. One potential solution to mitigate such risks involves adopting methods similar to anti-dreambooth~\cite{van2023anti}, which introduce subtle noise perturbations to the published images to mislead the customization process. Additionally, applying unseen watermarking to the generated video could deter misuse and prevent them from being used without proper recognition.

\end{document}